# Fast ground-to-air transition with avian-inspired multifunctional legs

Won Dong Shin[1]✉, Hoang-Vu Phan[1], Monica A. Daley[2], Auke J. Ijspeert[3], and Dario Floreano[1]

[1]Laboratory of Intelligent Systems, École Polytechnique Fédérale de Lausanne, Lausanne CH1015, Switzerland
[2]Neuromechanics Lab, University of California, Irvine, Irvine, CA 92697, USA
[3]Biorobotics Laboratory, École Polytechnique Fédérale de Lausanne, Lausanne CH1015, Switzerland
✉Corresponding author's email: wdshin123@gmail.com; won.shin@epfl.ch

**Abstract**

Most birds can navigate seamlessly between aerial and terrestrial environments. Whereas the forelimbs evolved into wings primarily for flight, the hindlimbs serve diverse functions such as walking, hopping, and leaping, and jumping take-off for transitions into flight[1]. These capabilities have inspired engineers to aim for similar multi-modality in aerial robots, expanding their range of applications across diverse environments. However, challenges remain in reproducing multi-modal locomotion, across gaits with distinct kinematics and propulsive characteristics, such as walking and jumping, while preserving lightweight mass for flight. This tradeoff between mechanical complexity and versatility[2] limits most existing aerial robots to only one additional locomotor mode[3–5]. Here, we overcome the complexity-versatility tradeoff with RAVEN (Robotic Avian-inspired Vehicle for multiple ENvironments), which uses its bird-inspired multi-functional legs to jump rapidly into flight, walk on ground and hop over obstacles and gaps similar to the multi-modal locomotion of birds. We show that jumping for take-off contributes substantially to initial flight take-off speed[6–9] and, remarkably, that it is more energy-efficient than solely propeller-based take-off. Our analysis suggests an important tradeoff in mass distribution between legs and body among birds adapted for different locomotor strategies, with greater investment in leg mass among terrestrial birds with multi-modal gait demands. Multi-functional robot legs expand opportunities to deploy traditional fixed-wing aircraft in complex terrains through autonomous take-offs and multi-modal gaits.

**Main text**

Birds power flight using wings, but multi-segmented legs are the key skeletal feature that enables birds to navigate almost all complex terrestrial terrains and transition between domains[1]. Birds adapt their gait kinematics to meet the demands of multiple locomotion modes[10,11] through varied activation of muscles to control motion of leg and foot segments at the hip, knee, ankle, and toe joints. In addition, birds can store elastic energy in the muscle-tendon systems of the leg during flexion and release it during extension for take-off[8]. These mechanisms allow birds not only to walk and jump to overcome gaps but also to rapidly accelerate to the velocity required for take-off[6,7,9]. However, powerful legs come at the cost of added weight during flight, and therefore a trade-off exists in investment of muscle mass for terrestrial versus aerial locomotor modes. As discussed below, this trade-off depends on the ecological niche and the size of the bird.

Multi-functional legs inspired by birds may also expand the locomotion capabilities of aerial robots. For example, fixed-wing drones display higher energetic efficiency and flight endurance compared to rotary-wing drones, but are restricted in ground mobility and take-off typically requires a runway or a launcher[12]. With bird-inspired legs, a winged drone could in principle walk, run, hop, and actively transition between ground and air by jumping for take-off—strategies commonly used by most birds[9,13,14]—and thus eliminate the need for a runway or launcher regardless of terrain conditions.

Developing such multi-modal aerial robots poses major challenges. First, adding more locomotion modes comes at the cost of increased mechanical complexity and mass, which may make flight inefficient or impossible. Consequently, most multi-modal aerial robots to date are limited to one additional locomotor mode or function[3–5,15,16] in addition to flight. Second, distinct gait kinematics and propulsive characteristics are required to generate distinct locomotor modes, such as walking and jumping. Walking requires a lower peak propulsive force to push the body forward in a rhythmic and continuous cycle. In contrast, jumping, especially for take-off, demands a much higher propulsive force over a short period of time to generate rapid and forceful leg extension that accelerates the body against gravity. Most bio-inspired bipedal walking platforms are too bulky and heavy for integration into drones that can jump and transition to flight[3,17,18]. Miniature robotic jumpers at the scale of small birds[19–22] that rely on passive and rapid release of elastic spring mechanisms to generate jumping forces are not suitable for typical walking gaits. Although such jumpers can navigate the ground by consecutive hopping[20,21]—a behavior observed in small arboreal birds[23]—hopping at the scale of large birds incurs high energetic cost, which may explain why larger birds walk[24,25]. Therefore, a new leg and foot design is required for winged drones to walk, hop, jump, and fly as birds do.

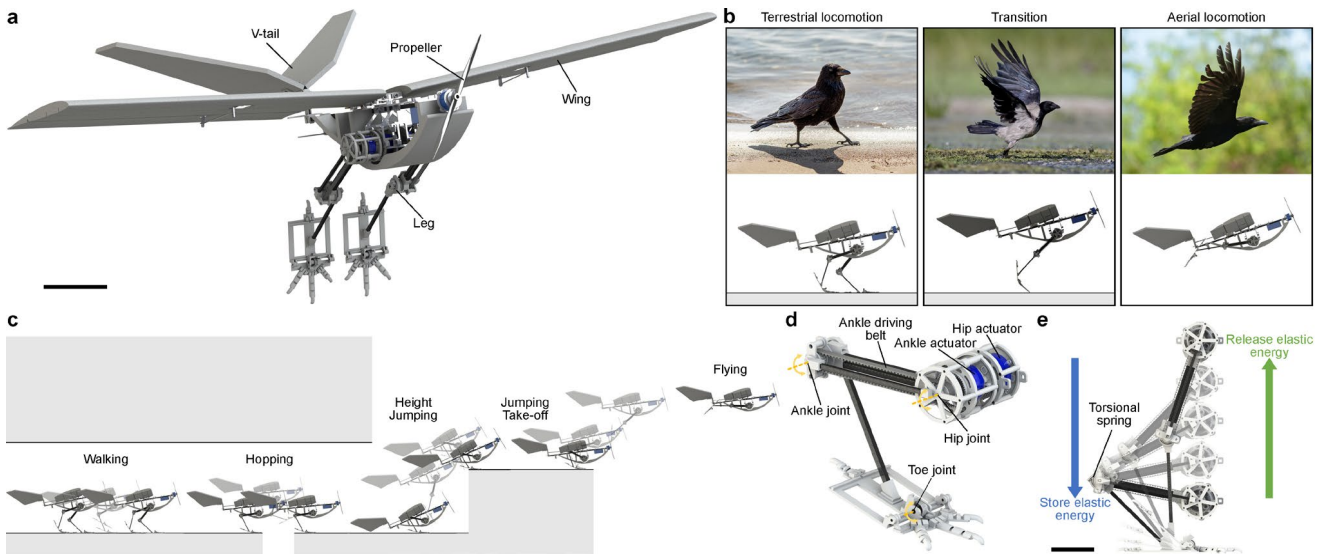

**Fig. 1| Avian-inspired robot design and capabilities. a**, RAVEN with a pair of legs, a pair of fixed wings, and a V-tail. RAVEN can deflect the wing ailerons and the V-tail to control flight. Scale bar, 10 cm. **b**, Similar to birds, RAVEN is capable of both terrestrial and aerial locomotion and active ground-to-air transition. **c**, When flying is not possible, the bipedal legs enable RAVEN to walk on the ground, hop to clear a gap, jump over an obstacle, and jump to take-off for flight. **d**, The hip and ankle joints of the leg mechanism are driven by two electromagnetic motors, and the toe joint of the foot is passively compliant with an embedded torsional spring. **e**, The ankle joint also has an embedded torsional spring, allowing storage and release of elastic energy during ankle flexion and extension, respectively. Scale bar, 5 cm.

**Bird-inspired biped mechanism**

Here, we present a bird-inspired biped mechanism that enables RAVEN (Robotic Avian-inspired Vehicle for multiple ENvironments) to walk, hop, jump onto an obstacle, and jump for take-off (Fig. 1a-c). Like the hindlimbs of birds, each robotic limb consists of a hip, ankle, and foot (Fig. 1d and Extended Data Fig. 1). Although birds have a more complex multi-segmented leg anatomy[26], replicating the full range of joints and degrees of freedom (DoFs) of a bird leg requires a high level of mechanical complexity that would substantially increase weight and control complexity. We therefore simplified the design into a two-segmented leg that rotates around a controllable hip joint connecting the leg to the body, and an ankle joint that increases the range of foot position[27] in the sagittal plane required for generating variable leg kinematics. The muscle mass of the bird hindlimb is distributed mostly in the upper leg (femur) near the hip joint, thus minimizing the moment of inertia and minimizing energetic cost[28]. To achieve a similar mass distribution, we located the driving electromagnetic actuators and gearboxes, which represent 64.5% of the leg's mass, at the hip joint (See Extended Data Table 1). We used a timing belt and pulleys to transfer the power output of the ankle actuator to the ankle joint (Fig. 1d).

A bird prepares for a jump by flexing its legs to store elastic energy in the distal joints, including the ankle[8]. By rapidly extending the legs from the flexed posture, the bird amplifies the peak power output by combining the energy stored in the tendon with power produced by the muscle to generate an explosive jump[8] (Supplementary Video 1). We capture the principle of energy storage and rapid release by embedding a torsional spring at the ankle joint (Fig. 1e) that can boost the instantaneous power output of the ankle actuator and improve jumping speed by 25% (Extended Data Fig. 2, and Supplementary Video 2).

Birds control motion of the multi-jointed foot by digital flexor and extensors[29,30] and motion of toe joints by the coupling of multi-articular muscles acting at the knee and ankle[17]. To reduce mechanical and control complexity, we designed a flat foot with a back claw and a passive elastic toe joint (Fig. 1d). We located the toe joint at a distance from the rigid intersection point between foot and the leg (see 'Foot design' in Method for detail) so that the drone can stand stably and passively (Extended Data Fig. 3) when the projection of its centre of mass (CoM) on the ground remains within the footprint created by the toe joint and the back claw. The toe joint includes also a torsional spring to passively recover from deflections induced by locomotion gaits and enable the foot to change its orientation while maintaining toe contact with the ground, thus helping to mitigate slippage and orienting the body at desired angles during terrestrial locomotion and ground to air transition (Fig. 1b, Extended Data Figs. 3 and 4, and Supplementary Video 1 and 3).

The positioning of bipedal legs in a winged drone can affect the location of the CoM, which plays a crucial role in the stability of both aerial and terrestrial locomotion. In predominantly terrestrial birds, the CoM is located cranial to the hip, close to the knee, which reduces moments about the hindlimb joints and contributes to reducing muscular activation and energetic cost[31]. In contrast, in birds that display both terrestrial and aerial locomotion or mostly aerial locomotion, the CoM is positioned in a more cranial and ventral location, which is in front of and below the center of aerodynamic forces (neutral point), thus facilitating passively stable flight[31,32]. We therefore placed the hips below and near the wing leading edge (Fig. 1b), resulting in a CoM location that meets both abovementioned conditions to support aerial and terrestrial locomotion.

**Ground-to-air transition**

To study the contribution of the bird-inspired legs to the acceleration into flight, we performed experiments on jumping take-offs of RAVEN with a weight of 620-gram, including the legs (Fig. 2). The leg motion and take-off speed (the speed at the instant when the feet leave the ground) were derived from simulation studies (see 'Dynamics modeling and simulation' in Methods, Extended Data Fig. 5, and Extended Data Table 2). The experimental results showed that by synchronously triggering the jumping legs and front propeller, the drone could transition to flight mode quickly and automatically (Fig. 2a,b, and Supplementary Video 1). The robot could achieve the desired take-off speed of approximately 2.4 ms$^{-1}$ in 0.17 s, which is comparable to birds of similar body mass[33] (see

'Design process' in Methods). We also conducted jumping tests without thrust from the propeller (Fig. 2b,c) and found that the legs could generate a speed of 2.2 ms$^{-1}$, thus contributing to about 91.7% of the required take-off speed (Fig. 2c; blue vertical dashed line at t = 0.17 s), similar to birds[9]. The jumping legs enabled the robot to start flight at a height of 0.4 m (about twice its body height), helping it clear potential ground obstacles.

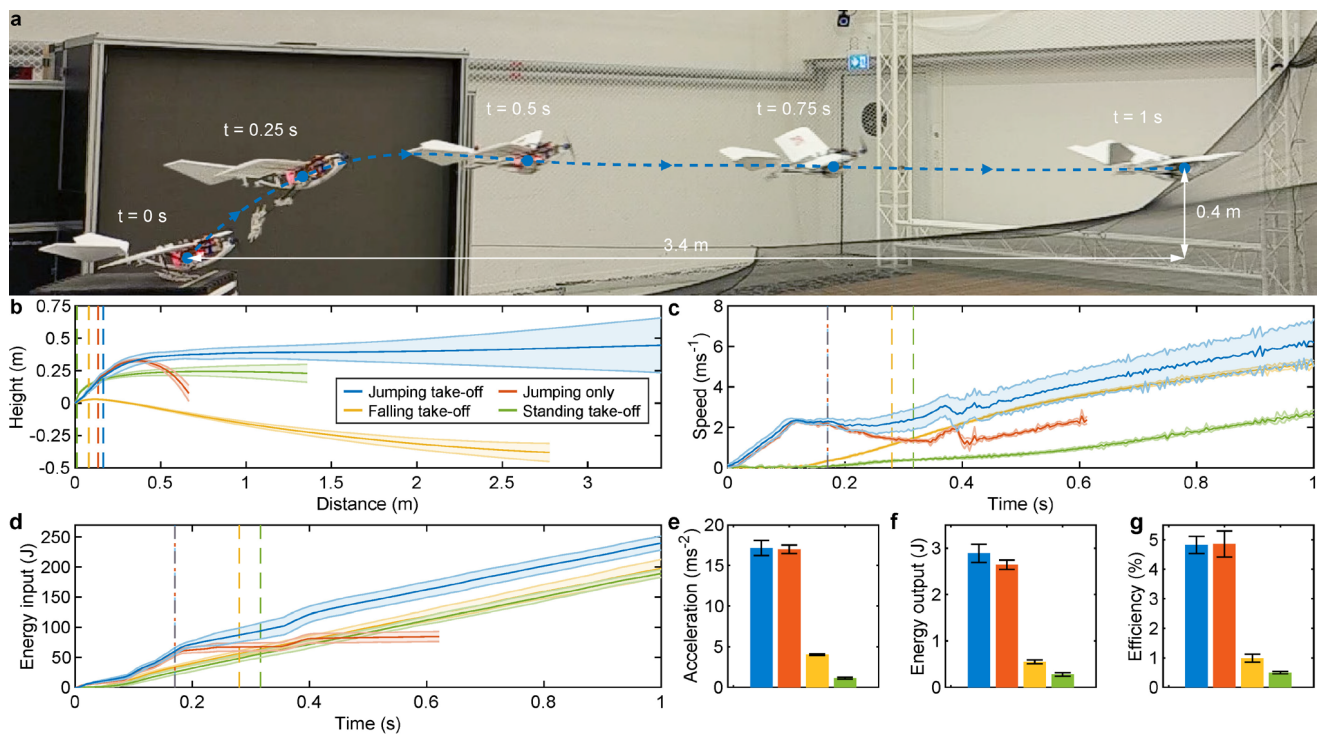

**Fig. 2| Take-off experiments. a**, RAVEN can jump to take off and can maintain stable level flight. **b-d**, Take-off trajectories (**b**), speeds (**c**), and energy inputs (**d**) of four different take-off strategies: Jumping take-off (blue), jumping only (red), falling take-off (yellow), and standing take-off (green). Solid line and shaded region represent the mean value and standard deviation, respectively, from five trials of each take-off strategy. The vertical dashed lines indicate the time of take-off. Jumping take-off, falling take-off, and standing take-off are shown in one-second period, and jumping ends at 0.62 s because the robot hit the ground. **e**, Average acceleration of the robot during the accelerating period before take-off (vertical dashed lines in **b-d**) obtained by taking derivative of speed in **c**. **f**, Total energy output is the sum of kinetic energy and potential energy at take-off. **g**, Efficiency is the ratio between the energy output and the energy input at take-off.

To examine the advantage of using jumping legs rather than using only the propeller for take-off, we conducted experiments on falling take-off and standing take-off where flight initiation relied solely on the propeller without engaging the legs. We then compared these take-off strategies with jumping take-off (Fig. 2b-d, 'Experiment setup' in Methods, Extended Data Fig. 6, and Supplementary Video 4). Falling take-off is a strategy that uses the propeller thrust to tilt RAVEN forward, fall from a height, and transition into flight. Standing take-off is a strategy that orients the body upwards to repurpose the propeller thrust and produce lift for take-off. To ensure consistency, we set the propeller thrust at the same level in all three cases. Although the falling take-off strategy enabled the drone to take off, it produced low take-off speeds of approximately 1.1 ms$^{-1}$ (Fig. 2c) and thus low lift, causing the drone to decrease flight altitude and thus convert potential energy into kinetic energy to increase speed for flight (Extended Data Fig. 6 and Supplementary Video 4). This take-off strategy is therefore applicable only when the drone and birds are at a height from the ground, such as on tree branches. Meanwhile, the standing take-off strategy generated the lowest take-off speed of about 0.4 ms$^{-1}$ (Fig. 2c) that, combined with the extremely high angle of attack of the wings, made it difficult to maintaining stable flight during the transition (Supplementary Video 4). In summary, the jumping take-off strategy allowed the drone to take off in a wider range of circumstances and at higher initial speed, thus leading to a more stable take-off.

We then compared the energy costs of the different take-off strategies (Fig. 2d and 'Energy consumption' in Methods). Although the energy cost of jumping take-off (60.1 J at the take-off point) was only slightly higher than the other two strategies (7.9% and 6.9% higher than standing take-off (55.7 J) and falling take-off (56.2 J), respectively, vertical dashed lines in Fig. 2d), it generated the highest take-off speed. To find out which take-off strategy is the most explosive (i.e. reaches the highest acceleration), the most energetic, and the most energy-efficient, we also compared the average acceleration before take-off (Fig. 2e), energy output (Fig. 2f), and energy efficiency at take-off (Fig. 2g) of the three take-off strategies, respectively (see 'Energy output and efficiency' in Methods). We found that jumping take-off was the best strategy in all three categories. Jumping take-off generated 15.1- and 4.3-fold higher average acceleration, and 10.4- and 5.3-fold higher energy output, compared to standing and falling take-offs, respectively. Remarkably, jumping take-off achieved 9.7- and 4.9-fold higher energy efficiency than standing and falling take-offs, respectively. These results demonstrate that the legs substantially contributed to the higher performance of jumping take-off (Fig. 2e-g; red columns). In summary, although jumping take-off requires slightly higher energy input, it is the most energy-efficient and fastest method to convert actuation energy to kinetic and potential energies for flight.

**Terrestrial locomotion**

To validate RAVEN's terrestrial locomotion abilities, we conducted experiments with three gait patterns: walking, hopping, and obstacle jumping (Fig. 3). We commanded the robot to generate a foot trajectory for each locomotion mode (see 'Foot trajectory generation' in Methods and Extended Data Fig. 7). The experiments showed that the drone could walk at a forward speed of 0.23 ms$^{-1}$ (Fig. 3a) and, although the gait was controlled in open loop, the (erect) drone could take several steps forward before becoming unstable and tumbling either nose-down or nose-up (Extended Data Fig. 8, and Supplementary Video 3). However, by resting the tail on the ground, the drone could passively stabilize its walking gait (Fig. 3a, and Supplementary Video 3). We then measured the ability to traverse a small gap by hopping. The drone could hop forward 26.6 cm (1.1× leg length) to clear an 11.5 cm distance gap and land stably (Fig. 3b, and Supplementary Video 5). Next, we assessed the ability to jump onto an obstacle by measuring vertical distance. The drone was able to jump up to 37.0 cm (1.5× leg length) and land on an obstacle with a height of 26 cm (Fig. 3c, and Supplementary Video 5). Combining the three gait modes allows the platform to accomplish the mission scenario of traversing a path with a low ceiling, hopping over a gap, and jumping onto an obstacle (Fig. 1c), which would be impossible for a fixed-wing drone on wheels[34,35] or equipped with jumping capability only[15,16,19]. A rotary wing drone of similar mass could tackle the mission scenario by flying over the obstacles, but it would likely be less energy efficient[36].

We resorted to the dimensionless Froude number and cost of transport (CoT) to compare the RAVEN speed and energy performance in horizontal gaits (walking and hopping) with those of other platforms at various scales. We included both terrestrial and multi-modal robots to assess the impact of including more operable environments in the performance of a robotic platform. As shown in Fig. 3g and h, multi-modal robots (blue shaded area) display lower Froude numbers but higher CoTs than terrestrial robots (red shaded area), suggesting that adding more locomotion modes reduces the robot's performance. Most multi-modal drones exhibited a higher CoT than terrestrial birds[37] and other walking and running animals[38] (Fig.3e). Compared to other multimodal drones, RAVEN displays relatively high Froude numbers and moderate CoTs, and unlike most existing multi-modal drones that typically only feature one terrestrial locomotion mode, RAVEN can perform three types of terrestrial locomotion (walking, hopping, and height jumping) as well as ground-to-air active transition.

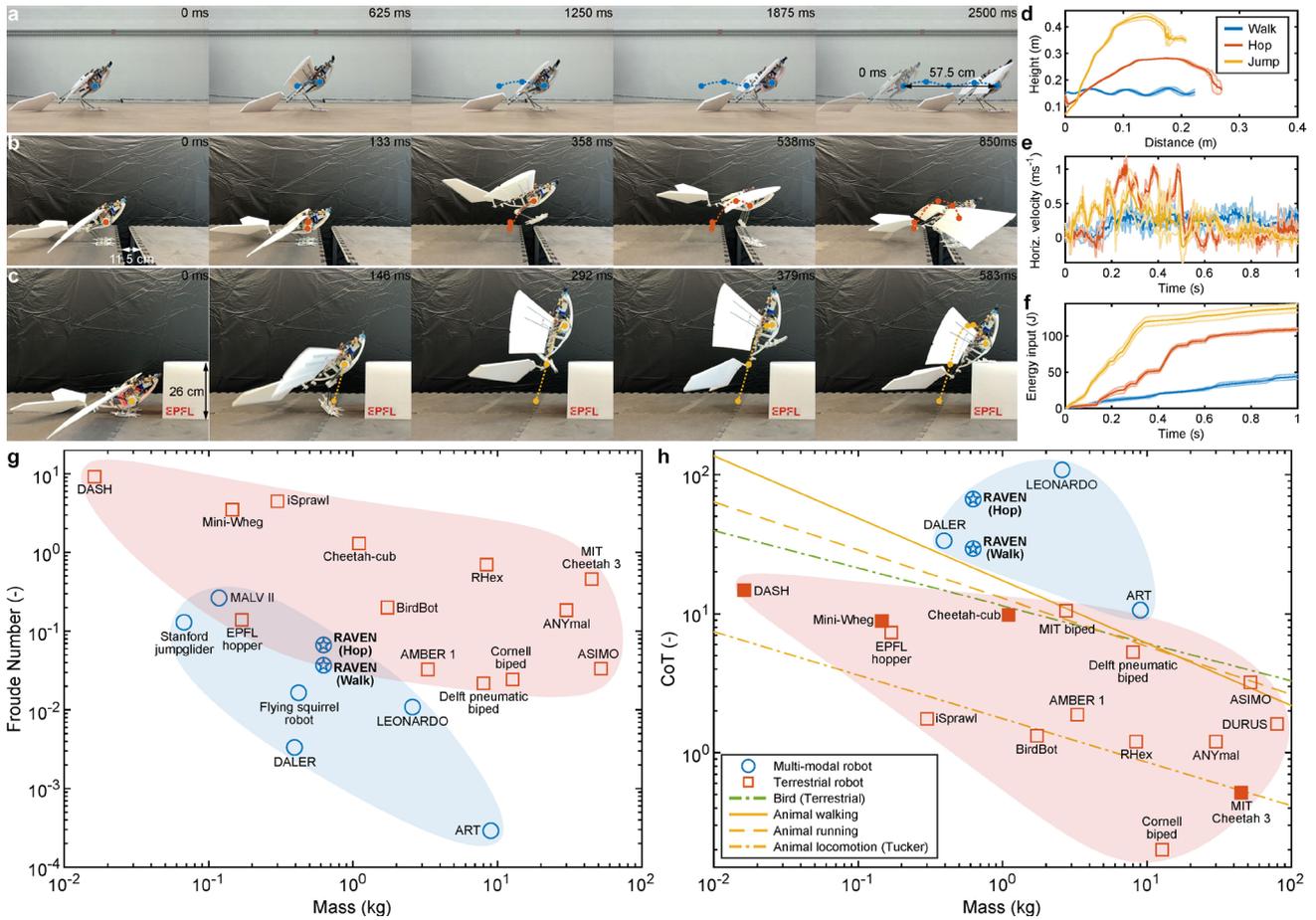

**Fig. 3| Terrestrial locomotion. a**, RAVEN is able to walk continuously at a speed of 0.23 ms$^{-1}$. **b**, hop over a 11.5 cm gap. **c**, and clear a vertical obstacle of 26 cm by jumping. The dashed lines in **a-c** represent the CoM trajectories. **d-f**, Trajectories (**d**), horizontal velocities (**e**), and energy inputs (**f**) of the three terrestrial locomotion modes. Solid line and shaded region are the mean and standard deviation (n = 5 trials) for each locomotion mode. **g-h**, The Froude numbers (**g**) and CoTs (**h**) of selected legged and multimodal aerial robots (see 'CoT and Froude number calculation' in Methods). Robots capable of both terrestrial and aerial locomotion are shown in blue, and robots that only operate on the ground are highlighted in red. This work is marked with stars. Filled markers in **h** indicate that only actuation energy is used for CoT calculation. The yellow solid and dashed lines in **h** are the net CoT trends of animals walking and running, respectively, from Rubenson et. al.[38]. The yellow dash-dotted line is the CoT of animals in general from Tucker[39]. The green dash-dotted line is the net CoT trend of running ground birds (galliformes) from Watson et. al.[37].

## Discussion

These results indicate that take-off initiated by jumping legs is a fast and energetically efficient strategy for flying systems that transition from the ground to the air, and reinforces the importance of jumping take-off as an everyday means for birds to transition between ground and air and to rapidly escape from predators[40]. Jumping take-off by RAVEN was enabled by avian-inspired legs that capture power-amplifying ankle joints that, as in birds[6–9], contribute to rapid acceleration for take-off speed, and passively compliant feet that allow take-off at the desired pitch angle. In addition, the leg design enabled adaptive locomotion gaits such as walking, hopping over a gap, and jumping onto an obstacle.

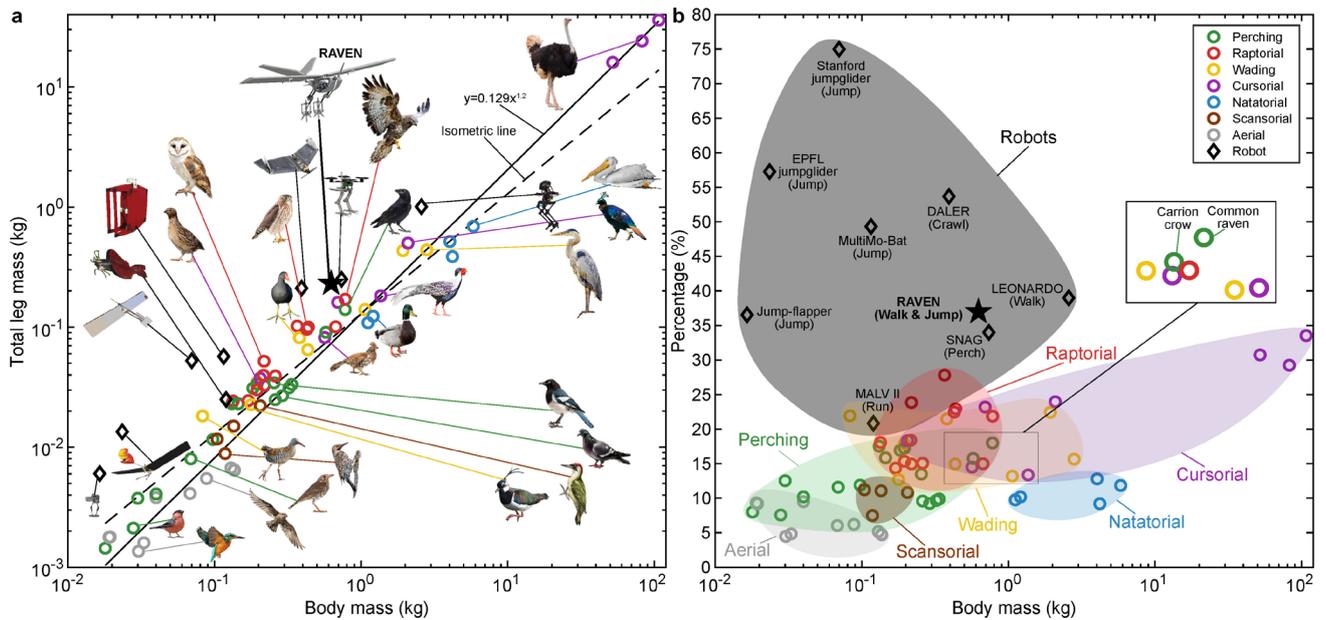

**Fig. 4| Leg mass budget of birds and selected multimodal aerial robots. a**, Total leg mass (for two legs) relative to total body mass among birds and multimodal robots. Circles are for birds, and diamonds are for aerial and terrestrial locomotion robots. RAVEN is highlighted with the filled star. The black solid line represents the trend line of the relationship between the total leg mass to total body mass in birds (see 'Bird data collection' in Methods). The dashed line is the line of isometry. Images of birds and robots are not drawn to scale. **b**, Total leg mass as a percentage of body mass. Bird groupings in **b** are based on Kilbourne[41].

Despite their utility for take-off, legs represent an extra payload in flight. Therefore, the partitioning of mass between legs and the rest of the body (the "leg mass budget") may pose an important evolutionary trade-off for birds. The optimal partitioning may depend on multiple factors such as body size, time spent in flying versus terrestrial gaits, and the frequency of ground to aerial transitions. The performance displayed by RAVEN suggests that even for mostly aerial birds, it is worth investing in leg mass to enable energy efficient and faster take-off that does not require starting from an elevated position. The performance benefits of legs are likely to be especially important for birds that frequently transition between air and ground. To test whether leg mass distribution relates to locomotion strategies, we collated data on leg mass budget for birds and aerial-terrestrial drones (Fig. 4). We compiled data from multiple sources and confirmed the finding that leg mass exhibits positive allometric scaling with body mass[28] (Fig. 4a). Additionally, we found that more terrestrial birds tend to "invest" more in leg mass (Fig. 4b). The most specialized aerial birds with limited use of ground-to-air transitions invest the least in leg mass. Scansorial birds (e.g. woodpeckers) invest more in leg mass than aerial birds, likely due to the use of legs for vertical climbing in arboreal environments. Natatorial (swimming) birds display a lower leg mass budget compared to other categories. In contrast, cursorial (adapted for running) and wading (walking through water) birds invest in higher leg mass compared to other bird groups, probably because they use legs for body weight support and propulsion on the ground. Raptorial (predatory) birds display high leg mass too, reflecting the use of legs to hunt prey. Perching birds are the most versatile, frequently switching between aerial, terrestrial, and arboreal modes of locomotion, and their leg mass budget is intermediate. Some individual species of perching birds, such as the common raven, have relatively heavier legs (Fig. 4b zoomed-in panel), consistent with terrestrial and arboreal legged locomotion and frequent use of ground to air transitions. These findings suggest functional trade-offs associated with leg muscle mass and suggests adaptive co-evolution of muscle mass with locomotor ecology among birds[42].

Constraints on power output of muscles[43] may prevent some large birds from generating sufficient jumping acceleration for take-off, leading to the use of a running take-off from land or water to get airborne (e.g. swans, albatrosses[44]). Robotic platforms might be less constrained by muscle power limits, and therefore jumping might remain viable for active ground-to-air transition for drones at mass scales beyond those observed in birds. Further research is required to scale up bird-inspired multifunctional legs to large drones capable of autonomously navigating between terrestrial and aerial environments. The design principles presented here pave the way for more versatile machines that, like animals, can use multiple locomotion modes while minimizing the energetic trade-offs associated with increased mechanical complexity.

**Online content**

Any methods, additional references, Nature Research reporting summaries, source data, extended data, supplementary information, acknowledgements, peer review information; details of author contributions and competing interests; and statements of data and code availability are available at https://doi.org/

## Methods

### Design process

**Mass and dimension.** We designed the robot on a scale similar to that of carrion crows, which are commonly found birds capable of walking, hopping, and taking off by jumping. With all onboard components, the robot weighs 600 gram, which is in the mass range of carrion crows (430-650 g[45]). The wingspan and body length are 100 cm and 50 cm, respectively, similar to those in carrion crows (84-100 cm and 44-51 cm[45]).

**Actuator selection.** Birds with body masses of 491 and 783 g take off at 1.85 and 3.21 ms$^{-1}$, respectively[33]. Using linear interpolation, we can calculate the desired take-off speed, $v_{\text{take-off}}$ = 2.5 ms$^{-1}$, for our 600-gram platform. Thus, the desired mechanical power at the moment of take-off, $P_{\text{take-off}}$, can be estimated as

$$P_{takeoff} = m_b g v_{take-off} \qquad (1)$$

where $m_b$ is the body mass, and $g$ is the gravitational acceleration (9.81 ms$^{-2}$). Thus, the desired mechanical power at the moment of take-off is 14.7 W. When selecting proper lightweight electric motors for the hip and ankle joints that generate the desired power output for taking off, we considered only 10% of the rated maximum power output of the motor as the available power because electric DC motors have speed-dependent power output, which is substantially low at a high-speed rotation. We also applied an extra safety factor of two for the costs from the transmission system such as the gearbox and timing belt. As a result, we could estimate the minimum required power output from the leg motors of 294 W. From this value, we selected four 18-gram T-Motor (AT2303 KV1500 Short Shaft, maximum power of 100 W) to drive the two legs.

**Gear reduction ratio.** We used a compact planetary gearbox to increase the torque output of the selected motor[46]. To obtain an appropriate reduction ratio of the gearbox, we simulated a dynamic model for the jumping take-off. In the simulation, we commanded the platform to jump and take off at 2.5 ms$^{-1}$ with zero angular velocity at an initial body angle of 10° (Extended Data Fig. 5). We found that the required maximum joint speed during the jumping take-off is 4384 °s$^{-1}$ (Extended Data Fig. 5). As the theoretical maximum speed of the motor is 99900 °s$^{-1}$, the gear reduction ratio should be less than 22.8. Using available off-the-shelf gears consisting of a 9-tooth sun gear, a compound planet gear having 10 and 32 teeth, and a 59-tooth ring gear, we could achieve a reduction gear ratio of 19.13.

**Foot design.** Round foot and flat foot are commonly used designs in robotic platforms[47,48]. Round foot provides unconstrained foot orientation with respect to the ground but does not provide a wide contact area for balancing[47,48]. Therefore, round foot design is widely used in robots having more than two legs because the multiple foot contact points enlarge the support polygon[49,50], the planar convex hull formed by the ground contact points. When the round foot design is used in bipedal robots, the presence of only two foot contact points makes it difficult to achieve passive stability, necessitating complex active control for balance. For this reason, bipedal platforms prefer to use flat feet to increase the area of support polygon. However, the flat foot design constrains the orientation of the lower limb with respect to the ground and, consequently, requires an extra joint on the foot where the lower limb and the foot are connected to each other[50,51]. This joint may require an additional actuator to control the angle of the joint.

Our foot design bypasses the problem of adding an extra actuator by offsetting the joint location toward the front. This offset creates two discrete foot surfaces (the palm surface and the front surface with toes) connecting together by the toe joint (Extended Data Fig. 1). If projection of the CoM on the ground is located behind the toe joint, the palm forms a support polygon with the other foot (either palm or toes depending on the orientation of the other foot) and support the body (Extended Data Fig. 3d). When the CoM is located in front of the toe joint, the lower limb starts to change its orientation with respect to the body, and the toes start to support the body. This CoM location dependency of the foot design allows our platform to walk with smooth lower limb orientation adjustment and to jump with desired pitch angle (Extended Data Figs. 3 and 4, and Supplementary Videos 2, 3 and 5). A torsional spring is installed to the toe joint to bring the toes back to the original position when the foot lifts off the ground. In addition, three front toes and the hallux attached to the back side of the palm are designed to enlarge the surface area of the foot for a better stability. (Extended Data Fig. 1).

The offset length of the toe joint is determined considering a smooth transition from the ground to the air during the jumping take-off. We placed the CoM projection slightly in front of the toe joint to allow forward tilting of the body during the jumping take-off.

**Dynamics modeling and simulation**
**Jumping take-off.** We used jumping take-off simulation to calculate the required gear reduction ratio and to obtain reference trajectories of hip and ankle for jumping take-off. We simulated the jumping take-off in a two-dimensional space assuming that the left and right legs are completely synchronized. We used Lagrange formulation with six generalized coordinates ($q \in \mathbb{R}^6$) defining the horizontal ($q_1$) and vertical ($q_2$) positions, pitch angle ($q_3$), upper limb angle ($q_4$), lower limb angle ($q_5$), and toe deflection angle ($q_6$) (Extended Data Fig. 5). We assumed non-slipping conditions between the foot and the ground during a jumping, as follows

$$\dot{\mathbf{p}}_c = \mathbf{J}_c \dot{\mathbf{q}} = 0 \qquad (2)$$

$$\ddot{\mathbf{p}}_c = \mathbf{J}_c \ddot{\mathbf{q}} + \dot{\mathbf{J}}_c \dot{\mathbf{q}} = 0 \tag{3}$$

where $\mathbf{p}_c$ is the constraint positions on the foot, $\mathbf{J}_c$ is the Jacobian matrix for the constraint position defined as

$$\mathbf{J}_c = \frac{\partial \mathbf{p}_c}{\partial \mathbf{q}} \tag{4}$$

The constraint positions are defined as

$$\mathbf{p}_c = \begin{bmatrix} \mathbf{p}_{toe} \\ \mathbf{p}_{claw} \end{bmatrix} \tag{5}$$

where

$$\mathbf{p}_{toe} = \begin{bmatrix} q_1 \\ q_2 \end{bmatrix} + \mathbf{R}(q_{4_{abs}}) \begin{bmatrix} l_1 \\ 0 \end{bmatrix} + \mathbf{R}(q_{5_{abs}}) \begin{bmatrix} l_2 \\ 0 \end{bmatrix} + \mathbf{R}(\theta_3) \begin{bmatrix} l_3 \\ 0 \end{bmatrix} \tag{6}$$

$$\mathbf{p}_{claw} = \mathbf{p}_{toe} + \mathbf{R}(q_{6_{abs}}) \begin{bmatrix} l_4 \\ 0 \end{bmatrix} \tag{7}$$

where $\mathbf{R} \in SO(2)$ is the rotation matrix defined as

$$\mathbf{R}(\theta) = \begin{bmatrix} \cos(\theta) & -\sin(\theta) \\ \sin(\theta) & \cos(\theta) \end{bmatrix} \tag{8}$$

and $l_1$ and $l_2$ are the upper limb and lower limb lengths, respectively, $l_3$ and $\theta_3$ are the offset length and angle of the toe joint, respectively, and $l_4$ is the toe length. The subscript $abs$ indicates the absolute angles with respect to the base frame. The parameters are depicted in Extended Data Fig. 5. The constraint positions change when the toe joint deflection reaches the maximum range and the body starts to rotate around the tip of the toe. In this case, the constraint position becomes

$$\mathbf{p}_c = \mathbf{p}_{claw} \tag{9}$$

The mentioned assumptions make the equation of motion during jumping take-off as follows[21]

$$\mathbf{M}(\mathbf{q})\ddot{\mathbf{q}} + \mathbf{C}(\mathbf{q}, \dot{\mathbf{q}})\dot{\mathbf{q}} + \mathbf{g}(\mathbf{q}) + \mathbf{J}_c^T(\mathbf{q})\mathbf{f}_c = \mathbf{S}^T \boldsymbol{\tau} + \mathbf{f}_{ext} \tag{10}$$

where $\mathbf{M}$ is the mass and inertia matrix, $\mathbf{C}$ is the centrifugal and Coriolis effect matrix, $\mathbf{g}$ is the gravitational force vector, $\mathbf{f}_c$ is the reaction force vector at the foot, $\mathbf{S}^T$ is the matrix that maps two by one input vector to generalized coordinates for a dimension match[52], $\boldsymbol{\tau}$ is the actuator input vector, and $\mathbf{f}_{ext}$ is the external force vector defined as

$$\mathbf{f}_{ext} = \mathbf{f}_p + \mathbf{f}_w + \mathbf{f}_t + \boldsymbol{\tau}_a + \boldsymbol{\tau}_t \tag{11}$$

where $\mathbf{f}_p$ is the force from the propeller, $\mathbf{f}_w$ and $\mathbf{f}_t$ are lift and drag forces of the wings and the tail respectively, and $\boldsymbol{\tau}_a$ and $\boldsymbol{\tau}_t$ are the spring torques on the ankle joint and the toe joint, respectively.

In order to achieve the desired take-off speed of 2.5 ms$^{-1}$ with zero angular velocity, we applied prioritized task-space control in the simulation. We prioritized the pitch, horizontal displacement, and vertical displacement in order. Then, we calculated the required acceleration values of generalized coordinates that satisfy our desired conditions of the three tasks. The acceleration of each prioritized

task is

$$\ddot{x}_i = \mathbf{J}_i\ddot{\mathbf{q}} + \dot{\mathbf{J}}_i\dot{\mathbf{q}} \qquad (12)$$

where the subscript $i$ indicates the $i^{th}$ task and $\mathbf{J}_i$ is the Jacobian matrix for the $i^{th}$ task. A lower $i$ value indicates a higher priority. We can reformulate equation (12) as

$$\mathbf{J}_i\ddot{\mathbf{q}} = b_i \qquad (12)$$

to make this equation in the form of linear equations where

$$b_i = \ddot{x}_i - \dot{\mathbf{J}}_i\dot{\mathbf{q}} \qquad (14)$$

We are interested in getting $\ddot{\mathbf{q}}$ when our desired conditions of the three tasks are satisfied. To find $\ddot{\mathbf{q}}$, we first need to solve for $\ddot{x}_i$, which is determined by our control choice. We used proportional–derivative (PD) control to achieve desired position and velocity, as follows

$$\ddot{x}_{i,d} = k_p(x_{i,d} - x_i) + k_d(\dot{x}_{i,d} - \dot{x}_i) \qquad (15)$$

where the subscript $d$ put on $x$ stands for 'desired', $k_p$ and $k_d$ are the position and velocity feedback constants, respectively. Equation (15) is used in equation (14) for $\ddot{x}_i$.

One way to put priority in the tasks is using the null-space projection[53]. To prevent a lower priority task from influencing a higher priority task, we introduce the augmented Jacobian matrix[53] defined as

$$\mathbf{J}_{A,i} = \begin{bmatrix} \mathbf{J}_1 \\ \vdots \\ \mathbf{J}_i \end{bmatrix} \qquad (16)$$

where the Jacobian matrices are stacked from the highest priority to the $i^{th}$ priority in order. When the null-space matrix of this augmented Jacobian matrix is multiplied with a task Jacobian matrix, we can prioritize tasks because the lower priority tasks are not able to influence higher priority tasks[53]. A null-space matrix of an augmented Jacobian matrix[53] is

$$\mathbf{N}_i = \mathbf{I}_k - \mathbf{J}_{A,i}^+\mathbf{J}_{A,i}, \quad \mathbf{N}_0 = \mathbf{I}_k \qquad (17)$$

where $\mathbf{N}_i$ is the null-space matrix for $i^{th}$ task, $\mathbf{I}$ is the identity matrix, $k$ is the number of generalized coordinates, the superscript '+' represents the pseudoinverse of a matrix[53]. We use damped least squares method[54] for the pseudoinverse matrix calculation defined as

$$\mathbf{A}^+ = \mathbf{A}^T(\lambda\mathbf{I} + \mathbf{A}\mathbf{A}^T)^{-1} \qquad (18)$$

where $\mathbf{A}$ is an arbitrary matrix, and $\lambda > 0$ is a damping constant. Recursively solving[52,53]

$$\ddot{\mathbf{q}}_i = \ddot{\mathbf{q}}_{i-1} + \mathbf{N}_i(\mathbf{J}_i\mathbf{N}_{i-1})^+(b_i - \mathbf{J}_i\ddot{\mathbf{q}}_{i-1}), \quad \ddot{\mathbf{q}}_0 = 0 \qquad (19)$$

gives the final solution, $\ddot{\mathbf{q}}_d = \ddot{\mathbf{q}}_3$.

In the next step, we calculate the required hip and ankle torque inputs that fulfill the $\ddot{\mathbf{q}}_d$ values under the non-slipping foot constraint. The QR decomposition[55] of the constraint Jacobian matrix

(equation (4)) is defined as

$$\mathbf{J}_c^T = \mathbf{Q}\begin{bmatrix}\mathbf{R}\\\mathbf{0}\end{bmatrix} \quad (20)$$

Multiplying $\mathbf{S}_u\mathbf{Q}^T$ to equation (10) without constraint and external force terms results in

$$\mathbf{S}_u\mathbf{Q}^T\mathbf{S}^T\tau = \mathbf{S}_u\mathbf{Q}^T(\mathbf{M}\ddot{\mathbf{q}}_d + \mathbf{C}\dot{\mathbf{q}} + \mathbf{g}) \quad (21)$$

where $\mathbf{S}_u = [\mathbf{0}_{3\times 2} \quad \mathbf{I}_3]$ that selects the unconstrained space columns of $\mathbf{Q}^T$.

We can rewrite equation (21) as

$$\tau = (\mathbf{S}_u\mathbf{Q}^T\mathbf{S}^T)^+\mathbf{S}_u\mathbf{Q}^T(\mathbf{M}\ddot{\mathbf{q}}_d + \mathbf{C}\dot{\mathbf{q}} + \mathbf{g}) \quad (22)$$

to obtain the hip and ankle torque values. The equation of motion (equation (10)) was solved for $\ddot{\mathbf{q}}$ using MATLAB 'ODE45' function.

**Flight.** After the jumping phase, the flight phase is modeled as

$$\mathbf{M}(\mathbf{q})\ddot{\mathbf{q}} + \mathbf{C}(\mathbf{q},\dot{\mathbf{q}})\dot{\mathbf{q}} + \mathbf{g}(\mathbf{q}) = \mathbf{S}^T\tau + \mathbf{f}_{ext} \quad (23)$$

where the foot constraint no longer exists. The hip and ankle joint torque inputs are given to hold the leg joints steadily in a fixed position.

**Simulation to reality gap**

In the simulation, we assume no limitation in computational power and data reading speed during the operation of the platform to calculate the actuator inputs. In reality, the computational power limitation of lightweight microcontrollers hinders from calculating required actuator inputs in real-time and causes delay in following input references. Due to these reasons, it is difficult to implement the exact control commands from the simulation to an actual platform. However, the simulation is still useful in many different aspects. First, the simulation was used to confirm possibility of generating jumping take-off with given physical design choices such as body mass, dimensions, actuators, etc. Second, the joint position and velocity trajectories from the simulation (Extended Data Fig. 5) were used as the references in the control of RAVEN. We converted the continuous joint velocity references from the simulation to discrete step functions (Extended Data Fig. 5g) considering the hardware limitation of the actuator and control system. We also delayed the activation of the hip joint (Extended Data Fig. 5f) because suddenly activating both hip and ankle joints at the same time caused belt skipping on the ankle joint. As a result, the angular position of the hip joint at the moment of take-off is less than that of the simulation (Extended Data Fig. 5f).

**Foot trajectory generation**

The foot trajectories were first calculated in MATLAB and then implemented on the actual platform. The foot trajectories during jumping take-off, walking, height jump and forward hop are present in

Extended Data Fig. 7. The jumping take-off consists of two phases, push-off phase and stretch-back phase, which is to remain the legs along the body during flight, as found in many birds. The push-off phase enables accelerating the body to achieve a desired take-off speed, and the stretch-back phase is to reduce drag during flight. The walking trajectory consists of a stance phase and a swing phase of the leg. The stance phase is the bottom half trajectory that pushes the body forward (Extended Data Fig. 7). The swing phase is the upper half trajectory that brings the foot back to the starting position without touching the ground. The height jump shares a similar push-off phase with the jumping take-off but at different initial body pitch angle. After the push-off phase, the legs are retracted toward the hip joint to prepare a landing. For the forward hop, the foot trajectory consists of five phases. It starts with a crouching phase to tilt the body pitch angle forward and adjust the jumping angle. The second phase is the push-off phase to accelerate the body. The third phase is for retracting the legs to prevent collision with the ground. In the fourth phase, the robot prepares landing by stretching the legs toward the ground. The final phase is the balancing phase in which the platform slowly places the CoM closer to the ground for a better stability (Extended Data Fig. 7).

**Hardware fabrication**

**Three-dimensional printing.** Custom designed parts are fabricated mostly using three-dimensional (3D) printing. All 3D printed parts that are mentioned in the following sections are printed using Ultimaker 5S with Tough Polylactic acid (Tough PLA).

**Gearbox.** Planetary gearboxes with a gear reduction ratio of 1:19.13 were custom designed and fabricated for both hip and ankle joints (Extended Data Fig. 1). We used off-the-shelf sun gears (brass) and planet gears (Polyoxymethylene), whereas the ring gears were 3D printed. A BLDC motor (T-Motor AT2303 KV1500 Short Shaft) were installed on a 3D printed flat base to drive the gearbox. Five round carbon tubes were also press-fitted to the flat base on the outer edge in a circular pattern with an equal spacing. 3D printed spacers were inserted to secure a space for the sun gear and three planet gears to mesh. Another flat 3D printed part with the ring gear was piled up on top of the spacers and press-fitted into the five round carbon tubes to close the gearbox. Two ball bearings were used for a smooth and well-aligned rotation of the gearbox.

**Leg and foot.** We aligned the rotation axes of two gearbox modules for the hip and ankle joints (Fig. 1d). The module closer to the body drives the hip joint to control the upper limb motion (Tibiotarsus), and the other module drives the ankle joint to control the lower limb (Tarsometatarsus) motion (Extended Data Fig. 1). The output of the ankle gearbox transmits power to the ankle joint using a Polyurethane timing belt (RS Components) and 3D-printed pulleys. The upper limb consists of three carbon fiber square rods (two 4 mm × 4 mm and one 3 mm × 3 mm) that connect the ankle gearbox and the ankle joint. The lower limb was made of a square carbon fiber rod (4 mm × 4mm) with one end connected to the ankle joint and the other end fixed to the foot. We also used torsional springs (Durovis) at the ankle joint and the toe joint. On the bottom of the foot, thin silicon (Ecoflex 20 from Smooth-On) pads were attached to increase friction.

**Chassis.** A 6 mm × 6mm square carbon rod was used as the backbone of the entire platform. 3D printed mounts for propeller motor, electric parts, legs, tail, and tail servomotor, were fixed to the backbone carbon rod using M2 metal screws and nuts.

**Propulsion, wings and tail.** We used a brushless motor (T-Motor AT2306 KV2300 Short Shaft), an electronic speed controller (T-Motor F30A 3-6S), and an 8-inch propeller (GWS 8040) for flight propulsion. We also used three servomotors (KST X08H) to actuate the wing ailerons and the tail elevator. Wings and tail made of Expanded Polypropylene (EPP) with a density of 20 $kgm^{-3}$ were fabricated using a CNC-Multitool CUT1620S foam cutter. Each wing was reinforced by two carbon fiber plates. We used SD7080 airfoil for the wings and a flat plate for the tail.

**Electronics**

We used a 3-cell battery (Gens Ace 700mAh) to power the robot. The battery was connected to a current sensor (ACS758KCB-150B-PSS-T) in order to measure the power consumption of the entire system. Five STM32 B-G431B-ESC1 boards were installed to the robot. One of the boards was used as the main controller, and the other four controllers were used to control the four electromagnetic motors that drive the left and right hip and ankle joints. The main controller commands the other four controllers via Controller Area Network (CAN) bus. We also used a secondary microcontroller (Seeed Studio XIAO nRF52840) to read the current data from the current sensor and record the data into a Secure Digital (SD) card through a SD card connector (SparkFun microSD Transflash Breakout). A six channel radio control (RC) receiver (FrSky RX6R) receives signals from a RC transmitter (FrSky Taranis X9D) and sends pulse width modulation (PWM) signals to the servomotors, the propeller motor controller, and the main controller. The main controller can send commands to the leg motor controllers based on the PWM signal input from the RC receiver. We used a constant 5 V voltage regulator (Murata OKI-78SR-5/1.5-W36-C) to power the main controller, the secondary microcontroller, the RC receiver, and the current sensor.

**Experimental setup**

For the jumping take-off experiments, we coupled the propeller thrust level and the leg activation by sharing the same PWM signal for thrust level and jumping initiation. By doing so, we made sure that the platform starts to jump with the identical thrust level for all trials. For the falling take-off and the standing take-off, the legs were commanded to hold a fixed position throughout the experiments. In the falling take-off, RAVEN tilts forward induced by the propeller thrust to fall from the ground and transition to flight. Standing take-off is a strategy in which RAVEN stands with a certain posture that directs the body in nearly vertical orientation. The take-off thus relies on only propeller thrust to lift up RAVEN and transition to flight. We set the throttle at 90% to achieve the maximum propeller thrust in all three take-off experiments for a fair comparison. For experiments that require leg activation including jumping take-off, height jump, forward hop, and walking, the platform generates pre-programmed foot trajectories.

**Robotic experiment data collection**

**Motion capture.** Experiments were conducted in a 10 m × 10 m × 8 m room equipped with a motion tracking system (26 Optitrack cameras). The position and attitude of the robot were recorded by the motion tracking system at a frequency of 240 Hz through four reflective markers attached to the platform. The velocity data were calculated by taking derivative of the position data. We also used high-speed cameras to record the experiments at 240 frames per second with a resolution of 1920 × 1080 pixels.

**Energy consumption.** The electrical energy consumption data were obtained from the current sensor. The analog current data is transmitted to an analog-to-digital converter (ADC) pin on the secondary controller. The maximum output of current sensor is 5 V, and the maximum input of the secondary controller is 3.3 V. Therefore, we installed a voltage divider consisting of a 2.7 kΩ resistor and a 4.7 kΩ resistor in series to scale down the analog data. The measured current data and timestamps were recorded and stored in a SD card. The power is then calculated by

$$P = VI \qquad (24)$$

where $V$ is the input voltage and $I$ is the current. The input voltage level of the battery was measured before and after each experiment, and the average value was used for the electrical power calculation. The electrical energy consumption is calculated as

$$E_{elec} = \int_0^{t_f} P(t)dt \qquad (25)$$

where $t_f$ denotes the time of each experiment.

**Energy output and efficiency.** We obtained energy output, $E_{out}$, which is sum of potential and kinetic energies, as follows

$$E_{out} = \frac{1}{2}m_b v^2 + m_b gh \qquad (26)$$

where $m_b$ is the mass of RAVEN, $v$ is the flight speed, $g$ is the gravitational acceleration, and $h$ is the altitude of the robot. The energy efficiency, $\eta$, was calculated as

$$\eta = \frac{E_{out}}{E_{elec}} \tag{27}$$

**Compilation of bird data**

The net CoT trend lines of walking and running of animals in Fig. 3h were taken from Rubenson et al.[38], and the net CoT trend line for terrestrial birds was from Watson et al.[37]. The CoT trend line of animals was taken from Tucker[39]. Rubenson et al.[38] includes non-minimum CoT in the data, but Tucker[39] only counts the minimum CoT. Considering this difference, the discrepancy is still large. The difference might come from different methods of collecting and calculating animal data. The total body mass and leg mass (for two legs) across birds spanning a large size were compiled from four publications [28,41,56,57]. From these data, we calculated the trend line (Fig. 4) using the MATLAB 'fit' function based on nonlinear least squares method with 'power1' option. The obtained equation is

$$m_l = 0.1289 m_b{}^{1.2} \tag{28}$$

with $r^2 = 0.99$ where $m_l$ is the total mass for two legs and $m_{body}$ is the total body mass.

**Other robotic platforms data compilation**

Froude numbers and CoT values were gathered from references[3,16,21,34,35,58–68]. Note that in some robots, we could not obtain either Froude number or the CoT due to the lack of available information. The leg and body masses of the robots were collected from references[3,4,15,16,34,35,66,69]. The leg mass includes any parts involved in the leg movement such as actuator, transmission, and the leg appendage itself. The leg mass of LEONARDO[3] was calculated based on the reported mass integration metric.

**CoT and Froude number calculation**

The Froude number was calculated as

$$FR = \frac{v_f{}^2}{gl} \tag{29}$$

where $v_f$ is the forward velocity and $l$ is the leg length defined as the vertical distance from the hip joint of the leg to the ground.

We estimated the CoT of RAVEN using the following equation:

$$CoT = \frac{E_{elec}}{m_b g d} \tag{30}$$

where $E_{elec}$ is the electrical energy spent during the operation and $d$ is the horizontal distance traveled.

## Data availability

All data are presented in the paper, Extended Data Figures, and Extended Data Tables.

## Code availability

MATLAB code for the jumping take-off simulation is available upon reasonable request.

**Acknowledgements**
This work was supported in part by the NCCR Robotics, a National Centre of Competence in Research, funded by the Swiss National Science Foundation under Grant 51NF40_185543 and in part by the European Union's Horizon 2020 research and innovation programme under Grant 871479 AERIAL-CORE.

**Author contributions**
W.D.S., H.-V.P., A.J.I. and D.F. conceived and designed the research, and analyzed the data. W.D.S. and H.-V.P. designed the experiment setups and wrote the manuscript. W.D.S. developed and simulated the jumping take-off model. W.D.S. designed, built, and programmed the robot with input from H.-V.P., and conducted experiments. W.D.S. generated the figures and videos. M.A.D. contributed to bird data analysis. A.J.I., M.A.D., and D.F. contributed to writing of the manuscript. All authors gave final approval for publication.

**Additional information**
**Correspondence and requests for materials** should be addressed to Won Dong Shin.

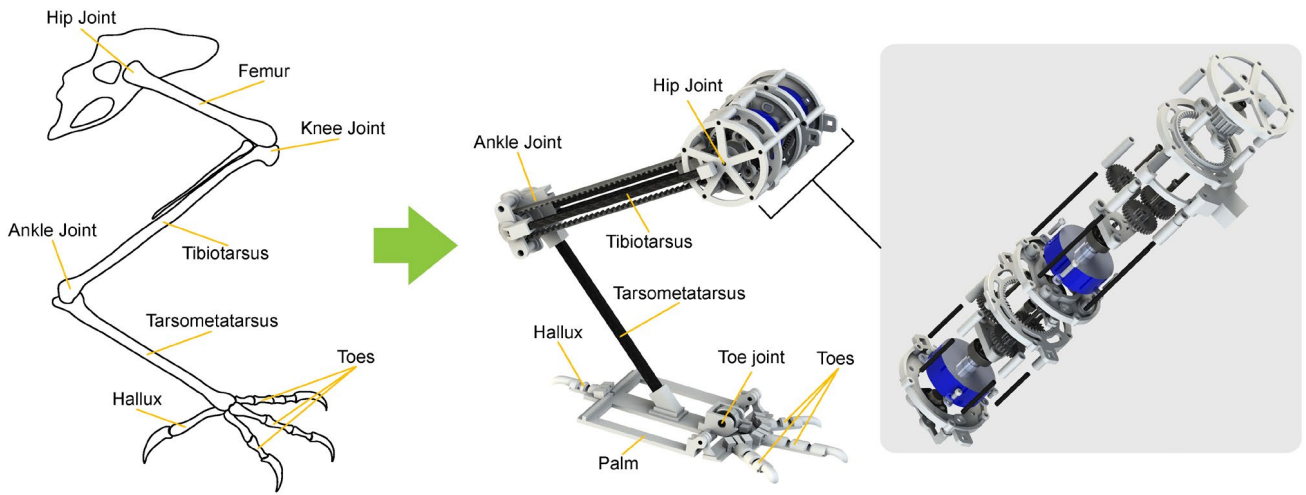

**Extended Data Fig. 1| Structure comparison of an avian leg and our leg design.** Our leg design omits the femur and knee joints to keep the actuated DoFs to two. The expanded view of the gearbox is presented.

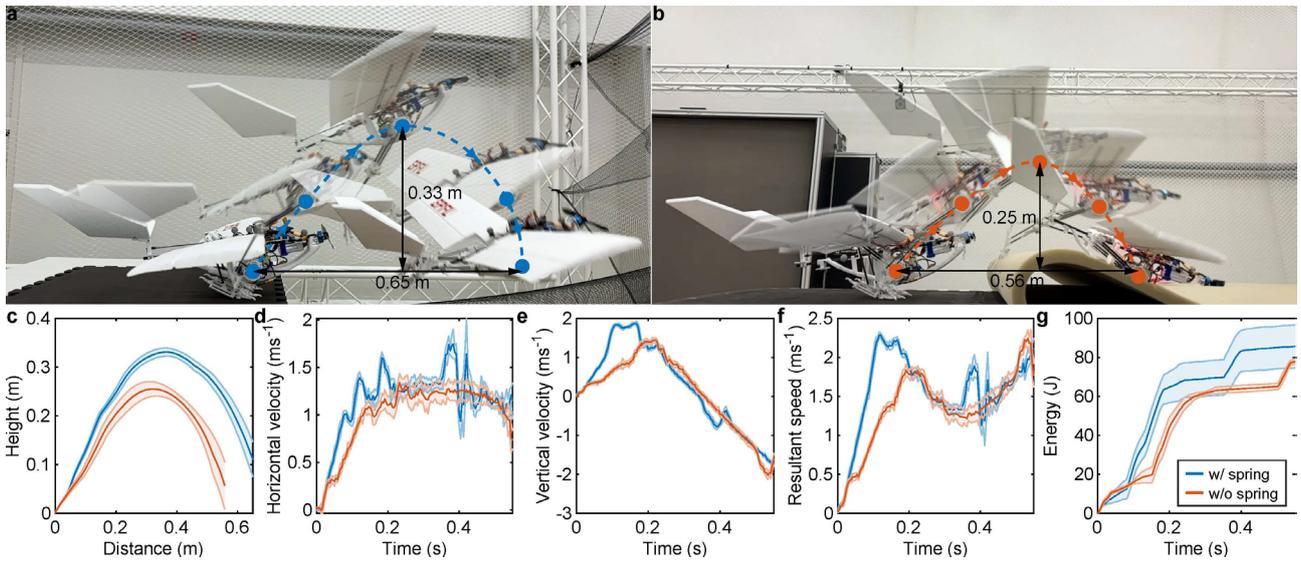

**Extended Data Fig. 2| Jumping comparison with and without ankle springs. a**, RAVEN jumping forward with springs on the ankle joints. **b**, RAVEN jumping forward without springs on the ankle joints. **c-g**, Experimental data comparing jumps with and without ankle springs on RAVEN. Five jumps are conducted for each condition. The solid lines are the mean values, and the shaded regions are the standard deviations.

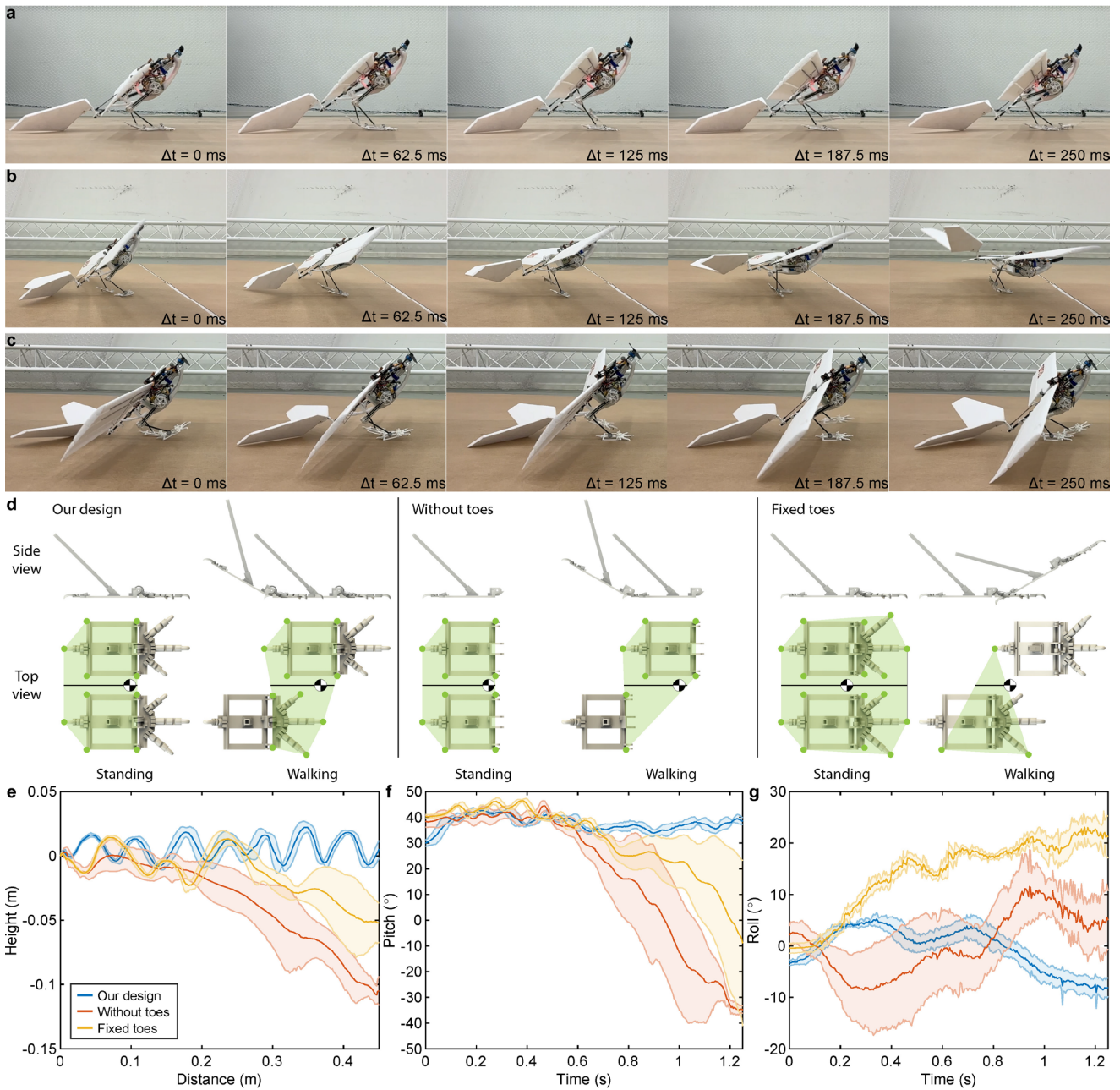

**Extended Data Fig. 3| Walking comparison with different foot designs. a-c**, Time sequential snapshots of walking with passively compliant toe joint (**a**), without toes (**b**), and fixed toe joint (**c**). The snapshots for **b** and **c** capture the moment of failures. The same walking trajectory is given to the three foot designs. **d**, Support polygon formations with different foot designs. The green circles indicate the contact extremities of the feet at the given posture (standing or walking). The green shaded regions are the support polygons formed by the contact extremities of the feet. **e-g**, Experimental data presenting the CoM trajectory, pitch angle, and roll angle. Five walking trials are conducted for each foot design. The solid lines are the mean values, and the shaded regions are the corresponding standard deviations. The trajectory data (**e**) are captured during the time period same as **f** and **g**.

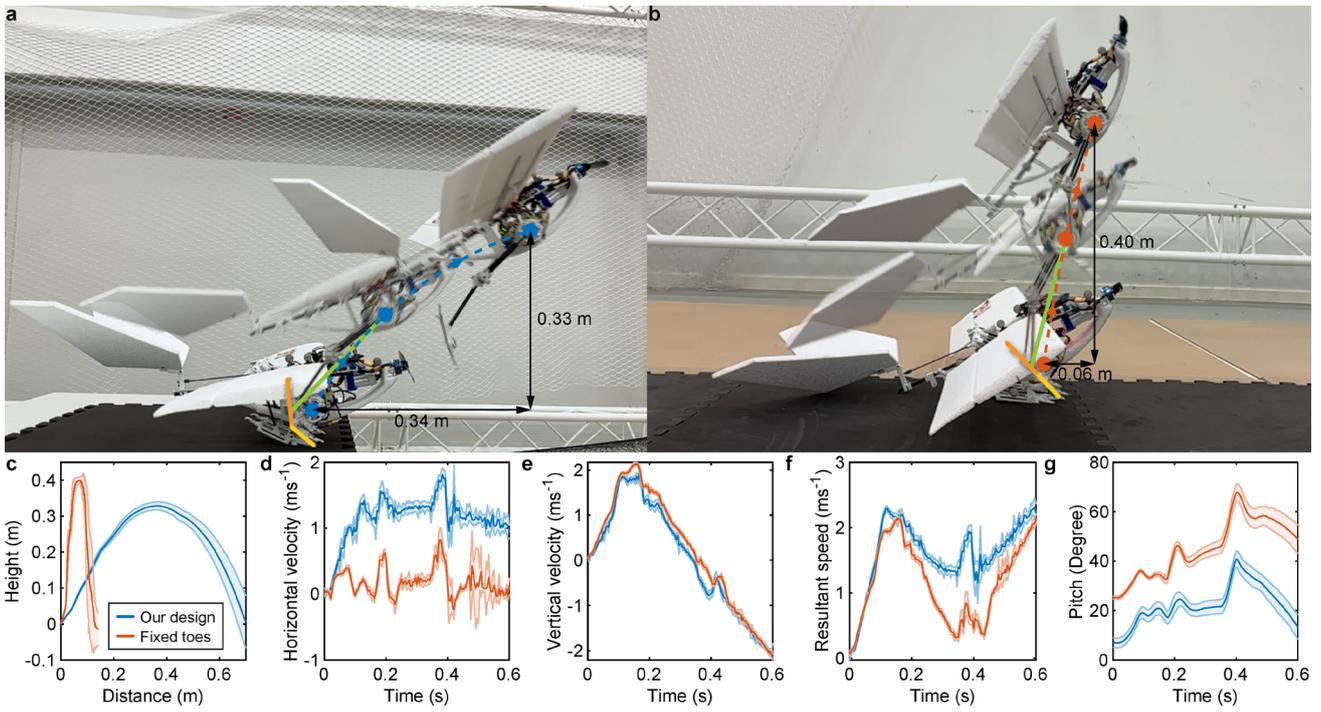

**Extended Data Fig. 4| Jumping comparison with different foot designs. a-b**, Jumping forward with the passively compliant toe joint (**a**) and fixed toe joint (**b**). The orange and yellow lines indicate the deflection between the palm and toes, respectively. The green line indicates the leg direction at the take-off. The same control command was given to RAVEN for both cases. **c-g**, Experimental data comparing jumps with passively compliant feet and fixed toe joint feet. Five jumps are conducted for each foot design. **d**, Experimental horizontal velocity data. **e**, Experimental vertical velocity data. **f**, Experimental resultant speed data. **g**, Experimental pitch angle data. The solid lines are the mean values, and the shaded regions are the corresponding standard deviations. The trajectory data (**c**) are captured during the time period same as **d-g**.

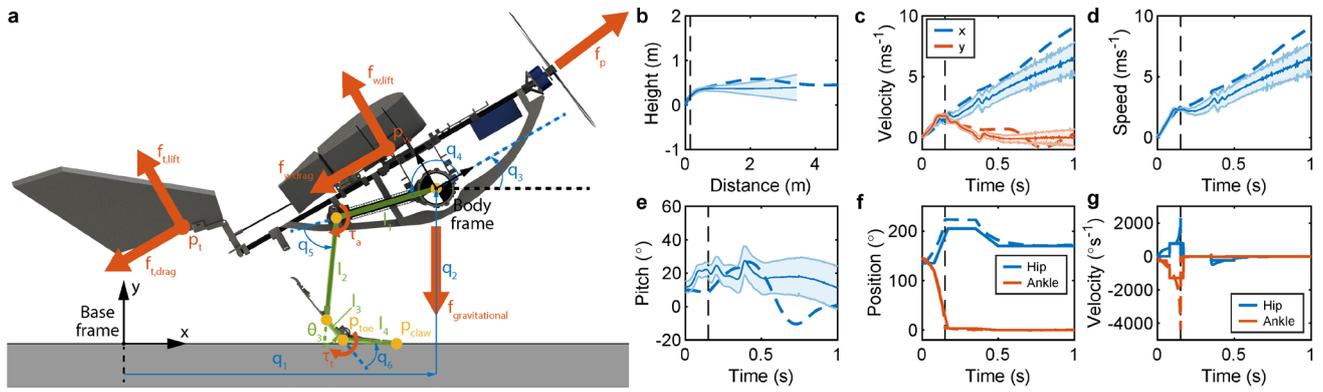

**Extended Data Fig. 5| Simulation and experimental results of jumping take-off. a**, Schematic of the model. The system is modeled with six generalized coordinates. The generalized coordinates are indicated with blue colour, and the constant parameters are coloured in green. The red coloured arrows are the external forces and torques. **b**, The CoM trajectory in 1 second time frame. The vertical dashed lines here and in **c-g** indicate the take-off moment. The coloured dashed lines in **b-g** are simulation results. The solid lines and shaded areas in **b-e** indicate the experimental mean values and standard deviations, respectively. **c**, Horizontal (x) and vertical velocity (y) of the CoM. **d**, The CoM resultant speed that combines the horizontal and vertical velocities. **e**, Pitch angle change over time. **f**, Hip and ankle joints positions. The solid lines in **f** and **g** are actual commands given to RAVEN **g**, Hip and ankle joints velocities.

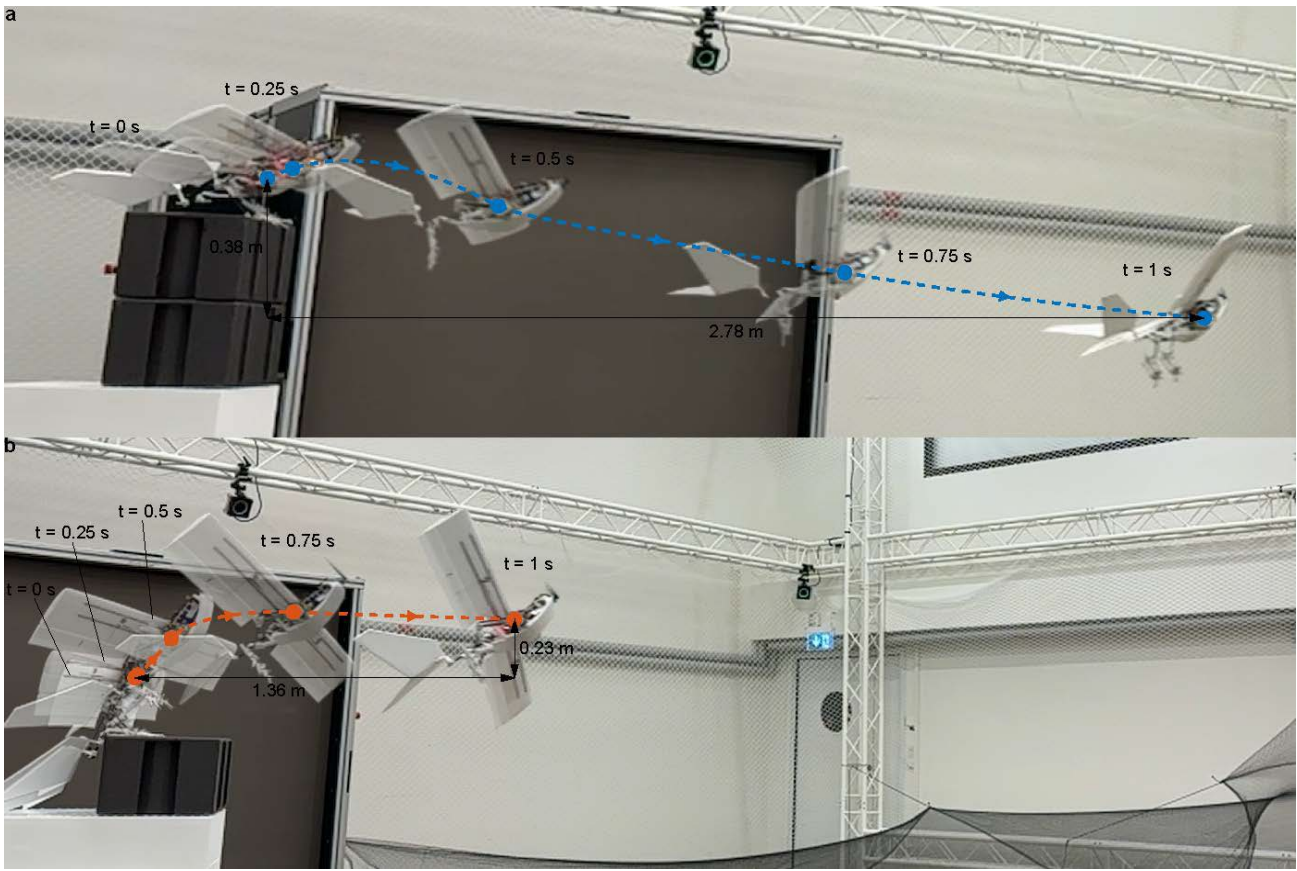

**Extended Data Fig. 6| Falling take-off and standing take-off time sequential snapshots. a**, Falling take-off time sequential snapshots. RAVEN lost 0.38 m in height but flew forward 2.78 m in 1 s. **b**, Standing take-off time sequential snapshots. RAVEN gained 0.23 m in height and moved forward 1.36 m forward in 1 s. The movement between t=0 s and t=0.25 s is subtle due to the near vertical direction of the thrust.

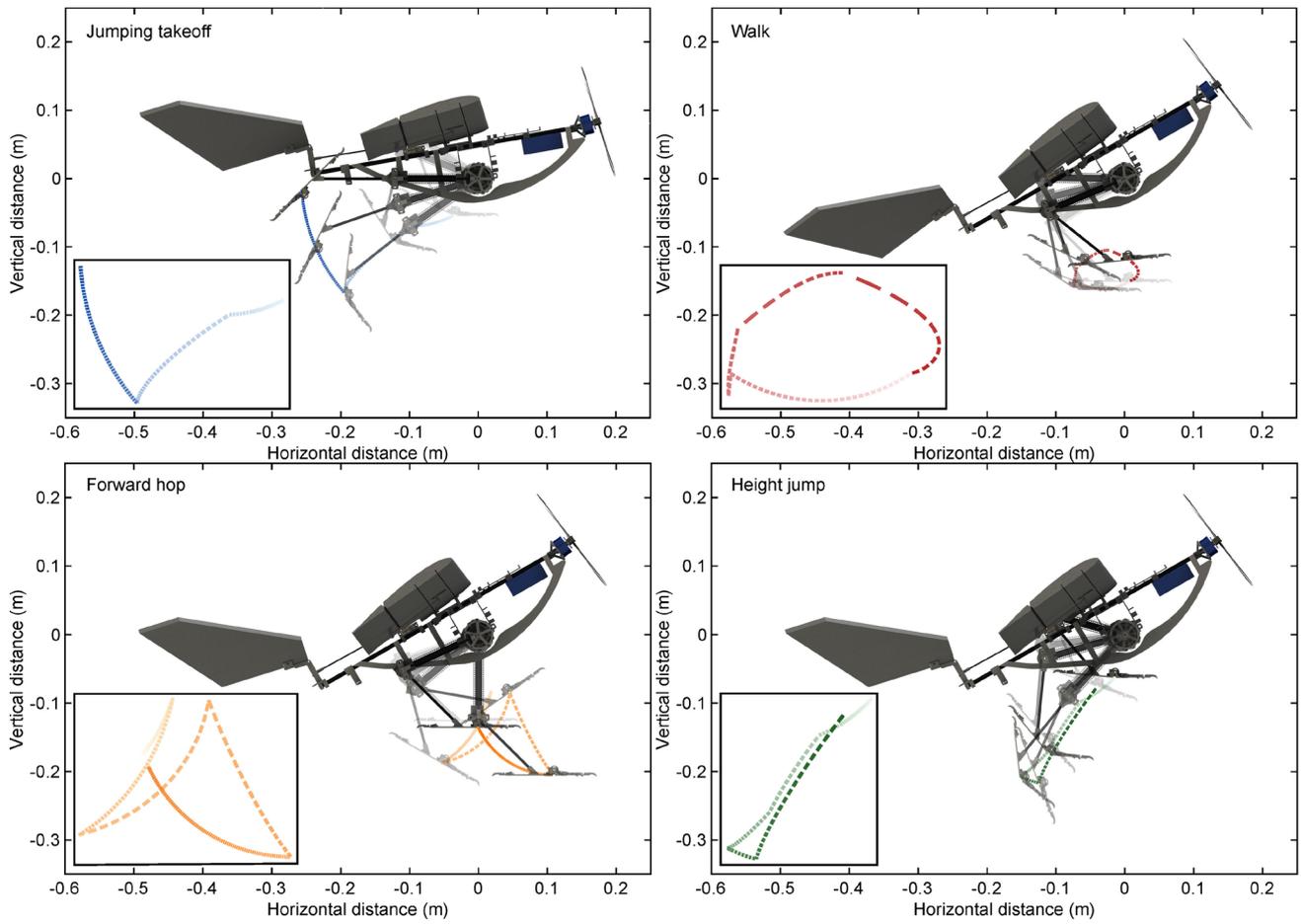

**Extended Data Fig. 7| Foot trajectories of the four types of locomotion.** The foot trajectories and the corresponding leg postures are presented. The sequence starts with the most transparent and ends with the opaquest leg posture and foot.

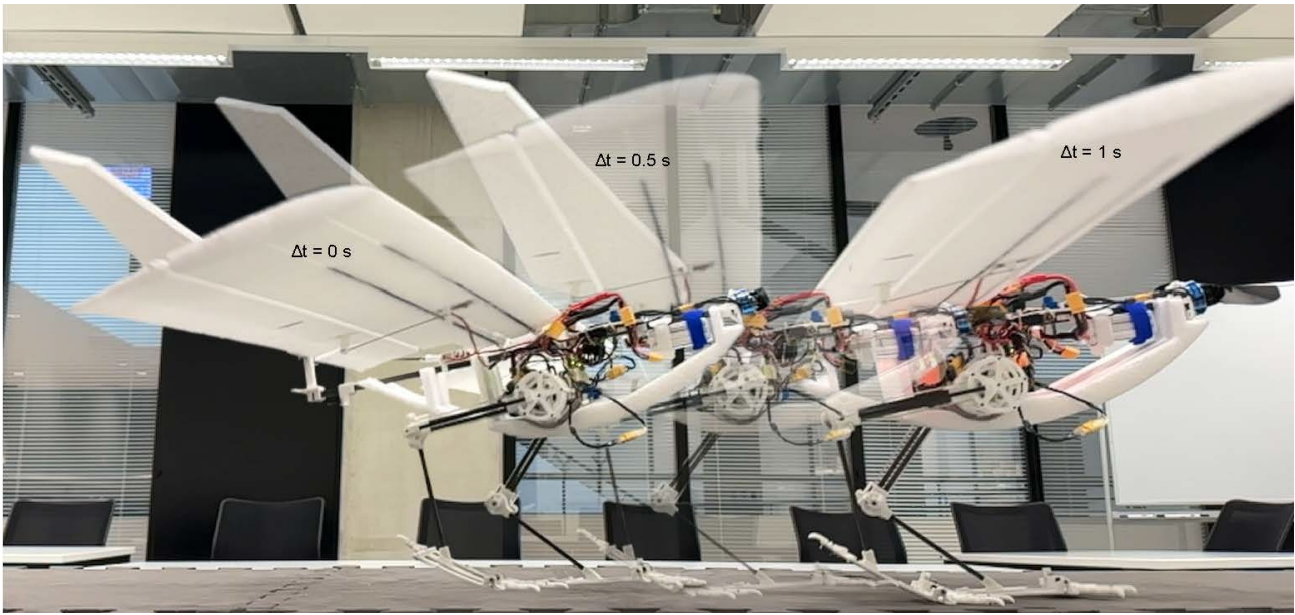

**Extended Data Fig. 8| Time sequential snapshots of erect walking.** RAVEN is capable of walking without the tail touching the ground but not continuously.

**Extended Data Table 1| Mass breakdown of RAVEN.**

| Part | Subpart | Mass (g) | Quantity | Subtotal (g) | Total (g) |
|---|---|---|---|---|---|
| Body stucture | Chassis (Swiss Composite CFK-Vierkantrohr 6x6 mm) | 10.80 | 1 | 10.80 | 60.57 |
| | Reflective markers | 9.52 | 1 | 9.52 | |
| | Protective foam | 9.06 | 1 | 9.06 | |
| | Electric components holding board | 4.31 | 1 | 4.31 | |
| | Leg holding frames | 3.80 | 2 | 7.60 | |
| | Leg rest | 3.00 | 1 | 3.00 | |
| | Wing connector | 5.26 | 1 | 5.26 | |
| | Propeller motor holder | 2.51 | 1 | 2.51 | |
| | Servo holder | 1.39 | 1 | 1.39 | |
| | Tail holder | 2.22 | 1 | 2.22 | |
| | Screw and nut | 0.35 | 14 | 4.90 | |
| Electric components on body | Battery (Gens Ace 700mAh 3s) | 52.48 | 1 | 52.48 | 189.36 |
| | Microcontroller (STM32 B-G431B-ESC1) | 9.36 | 1 | 9.36 | |
| | Motor controller (STM32 B-G431B-ESC1) | 9.36 | 4 | 37.44 | |
| | ESC (T-Motor F30A 3-6S) | 10.01 | 1 | 10.01 | |
| | RC Receiver (FrSky RX6R) | 4.00 | 1 | 4.00 | |
| | Current sensor (ACS758KCB-150B-PSS-T) | 4.67 | 1 | 4.67 | |
| | SD card reader (SparkFun microSD Transflash Breakout) + SD card | 2.72 | 1 | 2.72 | |
| | Secondary microcontroller (Seeed Studio XIAO nRF52840) | 2.40 | 1 | 2.40 | |
| | 5V Voltage converter (Murata OKI-78SR-5/1.5-W36-C) | 2.00 | 1 | 2.00 | |
| | Wires, pins, and connetors | 64.28 | 1 | 64.28 | |
| Aerodynamic surfaces | Wing | 30.80 | 2 | 61.60 | 143.21 |
| | Tail | 18.84 | 1 | 18.84 | |
| | Servo motor (KST X08H) | 8.94 | 3 | 26.82 | |
| | Motor (T-Motor AT2306 KV2300 Short Shaft) + connector | 30.73 | 1 | 30.73 | |
| | Propeller (GWS 8040) | 5.22 | 1 | 5.22 | |
| Legs | Motor (T-Motor AT2303 KV1500 Short Shaft) + connector | 19.80 | 4 | 79.20 | 230.04 |
| | Gearbox | 17.30 | 4 | 69.20 | |
| | Timing belt (Contitech Syncroflex 6/T2.5/285 SS) | 1.94 | 2 | 3.88 | |
| | Ankle spring (Durovis 16124) | 5.75 | 2 | 11.50 | |
| | Toe spring (Durovis 14081) | 1.52 | 2 | 3.04 | |
| | Upper limb | 9.71 | 2 | 19.42 | |
| | Lower limb | 4.43 | 2 | 8.86 | |
| | Foot | 17.47 | 2 | 34.94 | |
| **Total robot mass (g)** | | | | | **623.18** |

**Extended Data Table 2| Simulation parameters.**

| Parameter | Value |
|---|---|
| Initial pitch angle | 10° |
| Initial hip angle | 135° |
| Initial ankle angle | 145° |
| Initial toe deflection angle | 25° |
| Body mass | 0.5 kg |
| Upper limb mass | 0.015 kg |
| Lower limb mass | 0.015 kg |
| Palm mass | 0.015 kg |
| Toe mass | 0.005 kg |
| Body length | 0.5 m |
| $l_1$ | 0.12 m |
| $l_2$ | 0.12 m |
| $l_3$ | 0.023 m |
| $l_4$ | 0.06 m |
| $θ_3$ | 45° |
| Air density | 1.225 kg/m$^3$ |
| Wing area | 0.18 m$^2$ |
| Tail area | 0.0864 m$^2$ |
| Wing angle offset | 7° |
| Tail angle offset | 1° |
| Thrust angle offset | 7° |
| Max thrust | 0.63 kg |
| Ankle spring constant | 3.207 Nmm/° |
| Toe spring constant | 6.249 Nmm/° |
| $P_{w,x}$ | -0.02 m |
| $P_{w,y}$ | 0.042 m |
| $P_{t,x}$ | -0.28 m |
| $P_{t,y}$ | 0.074 m |